\documentclass[lettersize,journal]{IEEEtran}
\pdfoutput=1
\usepackage{amsmath,amsfonts}
\usepackage{algorithmic}
\usepackage{algorithm}
\usepackage{array}
\usepackage[caption=false,font=normalsize,labelfont=sf,textfont=sf]{subfig}
\usepackage{textcomp}
\usepackage{stfloats}
\usepackage{url}
\usepackage{verbatim}
\usepackage{graphicx}
\usepackage{cite}
\usepackage{threeparttable}
\usepackage{color}

\begin{document}

\title{A Coarse to Fine Framework for Object Detection in High Resolution Image}

\author{Jinyan Liu,~\IEEEmembership{Student Member,~IEEE,}
	Jie Chen,~\IEEEmembership{Senior Member,~IEEE,}
\thanks{The authors are with the School of Marine Science and Technology, Northwestern Polytechnical University, Xi’an 710072, China, and also with the Key Laboratory of Ocean Acoustics and Sensing, Ministry of Industry and Information Technology, Xi’an 710072, Shaanxi, China (e-mail: jinyanliu@mail.nwpu.edu.cn; dr.jie.chen@ieee.org)}
\thanks{Manuscript received April 19, 2021; revised August 16, 2021.}}

\markboth{Journal of \LaTeX\ Class Files,~Vol.~14, No.~8, August~2021}%
{Shell \MakeLowercase{\textit{et al.}}: A Sample Article Using IEEEtran.cls for IEEE Journals}


\maketitle

\begin{abstract}
Object detection is a fundamental problem in computer vision, aiming at locating and classifying objects in image. Although current devices can easily take very high-resolution images, current approaches of object detection seldom consider detecting tiny object or the large scale variance problem in high resolution images. In this paper, we introduce a simple yet efficient approach that improves accuracy of object detection especially for small objects and large scale variance scene while reducing the computational cost in high resolution image. Inspired by observing that overall detection accuracy is reduced if the image is properly down-sampled but the recall rate is not significantly reduced. Besides, small objects can be better detected by inputting high-resolution images even if using lightweight detector. We propose a cluster-based coarse-to-fine object detection framework to enhance the performance for detecting small objects while ensure the accuracy of large objects in high-resolution images. For the first stage, we perform coarse detection on the down-sampled image and center localization of small objects by lightweight detector on high-resolution image, and then obtains image chips based on cluster region generation method by coarse detection and center localization results, and further sends chips to the second stage detector for fine detection. Finally, we merge the coarse detection and fine detection results. Our approach can make good use of the sparsity of the objects and the information in high-resolution image, thereby making the detection more efficient. Experiment results show that our proposed approach achieves promising performance compared with other state-of-the-art detectors.
\end{abstract}


\section{Introduction}
\IEEEPARstart{O}{bject} detection is one of the most fundamental and important tasks in computer vision, which has attracted extensive attention over the past few years, for its ability to locate and classify objects in image\cite{survey1,survey2}. Object detection forms the basis of many other computer vision tasks, such as object tracking\cite{ref2}, instance segmentation\cite{ref3, mrcnn}, action recognition\cite{ref5}, scene understanding\cite{ref6}. Benefiting from the development of deep-learning based architectures, object detection has been more widely used in real world applications than before, such as autonomous driving\cite{survey1}, drone scene analysis\cite{visdrone}, and surveillance\cite{tinyperson}. 

Depending on whether using pre-defined proposals, the existing methods can roughly be categorized into two streams: anchor-based and anchor-free. The anchor-based detectors sample discrete bins from the bounding box space and refine the bounding boxes of objects accordingly. Anchors can be taken as regression references and classification candidates to infer proposals in multi-stage detectors, such as R-CNN\cite{rcnn}, Faster-RCNN\cite{frcnn} and Cascade-RCNN\cite{ref9}. Anchors can also be taken as predefined boxes for final regression in single-stage methods such as SSD\cite{ssd}, YOLO\cite{yolo}, and RetinaNet\cite{retinanet}. Anchor-free detectors such as FCOS\cite{fcos}, CenterNet\cite{centernet}, and GFL v1\cite{gflv1}, avoid complicated computation related to anchor boxes and bypass the settings of corresponding prior hyper-parameter.

These state-of-the-art approaches achieve good performance except for the small objects on natural image datasets with relatively low resolution, e.g., MS COCO\cite{coco} dataset (about 600$\times$400). To improve the performance on small objects in MS COCO\cite{coco} dataset, lots of methods including FPN\cite{fpn}, PANet\cite{panet}, BiFPN\cite{bifpn} adopt multi-scale feature learning approach. Designing better training methods is another approach, SNIP\cite{snip} and SNIPER\cite{sniper} improve accuracy by selecting train objects within a certain scale range. Perceptual GAN\cite{pergan} attempts to apply generative adversarial networks (GAN) to improve performance of small object detection. 

Different from MS COCO\cite{coco}, many other datasets that contain images suffering from extreme resolution and other distribution of objects' scale or position. For example, 
CityPersons\cite{citypersons} dataset contains images with a resolution up to 2048$\times$1024 and most of the objects in images are sparsely and sporadic as shown in Fig.1 (a), more than 35\% objects with a size smaller than 32$\times$32 pixels and the scale variance can be very large. TinyPerson\cite{tinyperson} dataset has many tiny objects (for example tiny persons less than 20$\times$20 pixels) in large-scale images (about 1920$\times$1080). The extremely small objects raise a grand challenge about feature representation while the massive and complex backgrounds increase the risks of false alarms. Besides, VisDrone\cite{visdrone}, UAVDT\cite{uavdt} and DOTA\cite{dota} datasets applying to unmanned aerial vehicle (UAV) vision contain images that parse large-scale scenes and quite small and dense objects that have extreme-scale variation.

In these cases, most of the existing state-of-the-art object detection methods encounter difficulties with high-resolution images, too small objects, and large-scale variance of objects. \cite{adap, tiles} divide an image into several small chips and then perform detection on each chip. In scale match method\cite{tinyperson} for TinyPerson\cite{tinyperson}, the original images are cut into some sub-images with overlapped parts because performing detection on large resolution images results in excess resource consumption during training and testing phases. But some objects may be divided into two parts in different chips resulting in detection precision of large objects decreasing. Besides, many sub-images become the pure backgrounds (no object in the image) due to image cutting, performing detection on them further leads to more time-consuming inference processing. \cite{tiles,clusdet,exploit,densitymap,penet} use sequential search strategy to detect small objects better in UAV vision images. The methods based on sequential search strategies use a coarse-to-fine framework that extracts interest regions from images and performs further detection. But most of them need to design specific modules or structures, thus leading to the methods' lack of universality. Some works avoid directly detecting tiny objects from high-resolution images. The VisDrone\cite{visdrone} dataset sets the ignored regions that contain small-scale objects in the distance to avoid the disadvantages in the test results. 

On the other hand, current devices can easily take very high-resolution images, even up to 5760$\times$3240 pixels. The higher the spatial resolution of the image in the same scene, the richer the detailed features.
The high-resolution images make the object detection accuracy higher and enable extremely small-scale objects at a further distance to be detected, which can be better applied to real world applications. However, current object detection methods seldom consider directly detecting tiny objects and solving large-scale variance problem in very high-resolution images to make full use of the information carried in images. Most existing state-of-the-art detectors encounter bad performance when detecting the origin or the down-sampled images. 
High-resolution input image results in problems of not only being out of memory of GPU but also slow convergence processing due to low mini-batch size, which brings severe challenges to hardware devices, especially for edge devices. And inputting the down-sampled high-resolution image will greatly reduce the average precision of the detection results, especially for small objects.

To deal with the issue mentioned above, in this paper, we propose a two-stage object detection method with coarse-to-fine philosophy. It almost makes full use of the high-resolution information in images and alleviates the problems of inference inefficiency and excessive memory consumption caused by the high-resolution input images. As illustrated in Figure \ref{fig:2}, our approach consists of five key steps including the center localization for small objects, the coarse detection, cluster region and chip generation, fine detection on chips, and results fusion. Firstly, the proposed framework conducts coarse object detection on the down-sampled image and obtains center locations of small objects in a high-resolution image by a lightweight detector at the first stage, and then filters the detected boxes and generates cluster regions based on the proposed cluster-based regions generation method. The cluster regions are cropped out from the high-resolution original image to see more details of small objects and fed to the chips generate module. Then, each chip is fed to the detector for fine detection. The final detection results are achieved by fusing results of both the chips and the global image. In summary, our paper has the following main contributions:
\begin{itemize}
\item We proposed a simple but effective coarse-to-fine object detection approach to improve the performance of relatively small-scale objects while maintaining the accuracy of large-scale objects in high-resolution images. 
\item We used a lightweight network to locate the center points of small objects by predicting the heat-map of the input image. And our approach can choose any existing state-of-the-art network without specific designs for coarse detection and fine detection stage to make the application more universal.
\item The proposed chips merging method can not only improve the detection accuracy and detection efficiency but also greatly reduce the memory consumption of the detector.
\item Our proposed framework can achieve 75.1 $AP_{50}$ on the CityPersons\cite{citypersons} dataset and 46.4 $AP_{50}$ on the TinyPerson\cite{tinyperson} dataset, which exceeds most existing state-of-the-art detection methods.
\end{itemize}
The rest of this paper is organized as follows. Section II briefly reviews the related works. In Section III, we describe the proposed approach in detail. Experimental results are shown in Section IV, followed by the conclusion in Section V.

\begin{figure}[htb]
	
	\begin{minipage}[b]{1.0\linewidth}
		\centering
		\centerline{\includegraphics[width=8.5cm]{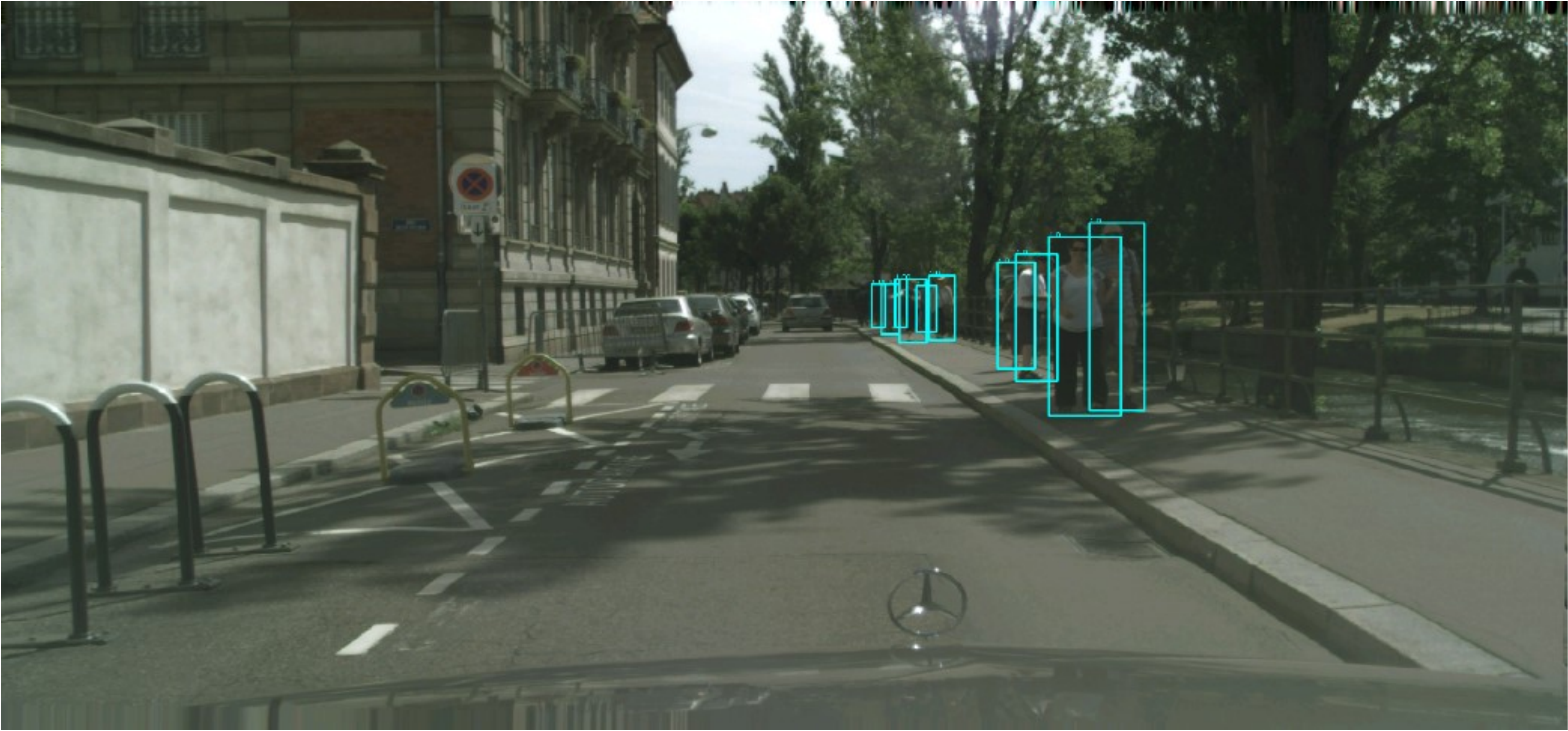}}
	\end{minipage}
	\caption{An image in CityPersons that objects in the image is sparse distributed. The blue box represents the ground truth boxes, the size of the image is 2048$\times$1024. It is very inefficient to feed such a high-resolution image directly to the detector.}
	\label{fig:1}
\end{figure}

\section{Related Work}
Object detection has been extensively explored in recent decades with a huge amount of literatures. In this section, we review convolutional neural networks (CNNs) based object detectors, small object detection and sequential search methods for object detection.
\subsection{CNN-based Object Detectors}
With the development of CNNs, the object detection methods have made a huge breakthrough. As a seminal anchor-based and region-based method, Faster-RCNN\cite{frcnn} separates object detection pipeline into two steps including proposals boxes extraction, and further classification and fine-tune regression of them. Anchors are taken as regression references and classification candidates to infer proposals in multi-stage detectors. On the contrary, the anchor-based but region-free detectors such as RetinaNet\cite{retinanet}, it performs detection without regions proposal and formulates detection as a regression task. Then, many anchor-free detector are proposed. FCOS\cite{fcos} regards all the locations inside the object bounding box as positives samples and then utilizes the centrality score to weight the positive and negative samples. Centernet\cite{centernet} defines the detection task as the integration of the center point of objects and the distance from it to the edge. TTFNet\cite{ttfnet}, as an improved version of Centernet\cite{centernet}, it uses Gaussian kernels to encode training samples for both localization and regression, and accelerates convergence of training processing via more supervised signals. Recently, the transformer-based encoder-decoder architecture has also been applied to object detection. DETR\cite{detr} views object detection as a direct set prediction problem and adopts transformer architecture as the backbone.

\subsection{Small Object Detection}
Small object detection on MS COCO\cite{coco} dataset has become a critical topic and attracted great attention from researchers. FPN\cite{fpn} is a major method to handle scale variance through feature pyramid, and it is extended to some variants, such as EFF-FPN\cite{efffpn} and AugFPN\cite{augfpn}. SNIP\cite{snip} and SNIPER\cite{sniper} use image pyramid to train and test detectors on the same scales and selectively back-propagates the gradients of objects with different sizes via a function of the image scales. Autofocus\cite{autofocus} adopts a coarse to fine approach and learns where to look in the next scale in the image pyramid. Dynamic zoom-in network\cite{dynamic} employs reinforcement learning to model long-term reward in terms of detection accuracy and computational cost, then dynamically select a sequence of regions for further detection at higher resolution. Perceptual GAN\cite{pergan} utilizes an adversarial network to boost the detection performance by narrowing the representation difference between small and large objects, and a super-resolution feature generator is trained with proper high-resolution object features for supervision. A similar idea appears in\cite{better} but considers the impacts of the receptive fields of various sizes as well. TinyPerson\cite{tinyperson} claims that scale mismatch between the data for network pre-training and detector learning incurs degradation and designs Scale Match to align object scales between different datasets.

\subsection{Sequential Search}
Sequential search refers to the methods that first determines the regions of interest in the image and then extracts these sub-regions from the image for further detection, which avoids processing the entire image and thus improve the detection accuracy and efficiency. Sequential search is usually used in object detection about drone or traffic sign images that contain a large amount of instances with very small size, because the common object detection methods will encounter trouble or become inefficient when dealing with these problems. Lu et al.\cite{adap} improved localization accuracy by adaptively focusing on sub-regions which likely to contain objects. 
Gao et al.\cite{dynamic} used a reinforcement learning strategy to adaptively zoom in the sub-regions for fine detection. There are also some works of sequential search for detecting small objects in high-resolution aerial images. Yang et al.\cite{clusdet} designed the ClusDet network and partitioned a high-resolution image into multiple small image chips for detection. Unel et al.\cite{tiles} split an image evenly with fixed size and verified the effect of sub-images for small object detection. zhang et al.\cite{exploit} paid attention to learn regions with low scores from a detector and gained performance by using better scoring strategy for regions with low score. DMNet\cite{densitymap} and CDMNet\cite{coarsedensity} introduced density map into aerial image object detection, where density map based cropping method is proposed to utilize spatial and context information between objects for improved detection performance. Tang et al.\cite{penet} adapatively generated clusters for different images in the dataset. In UFPMP-Det\cite{ufpmp}, sub-regions given by a coarse detector are initially merged through clustering to suppress background and  sub-regions are subsequently packed into a mosaic for a single inference. Besides, this method addresses the more serious confusion between inter-class similarities and intraclass variations of instances. Liu et al.\cite{focusfirst} presented a traffic sign detection method with a coarse-to-fine framework, which sequentially detects the objects in grid-level and image-level. 

\section{Proposed Method}
\subsection{Overview}
\begin{figure*}[htb]
	\begin{minipage}[b]{1.0\linewidth}
		\centering
		\includegraphics[width=17cm]{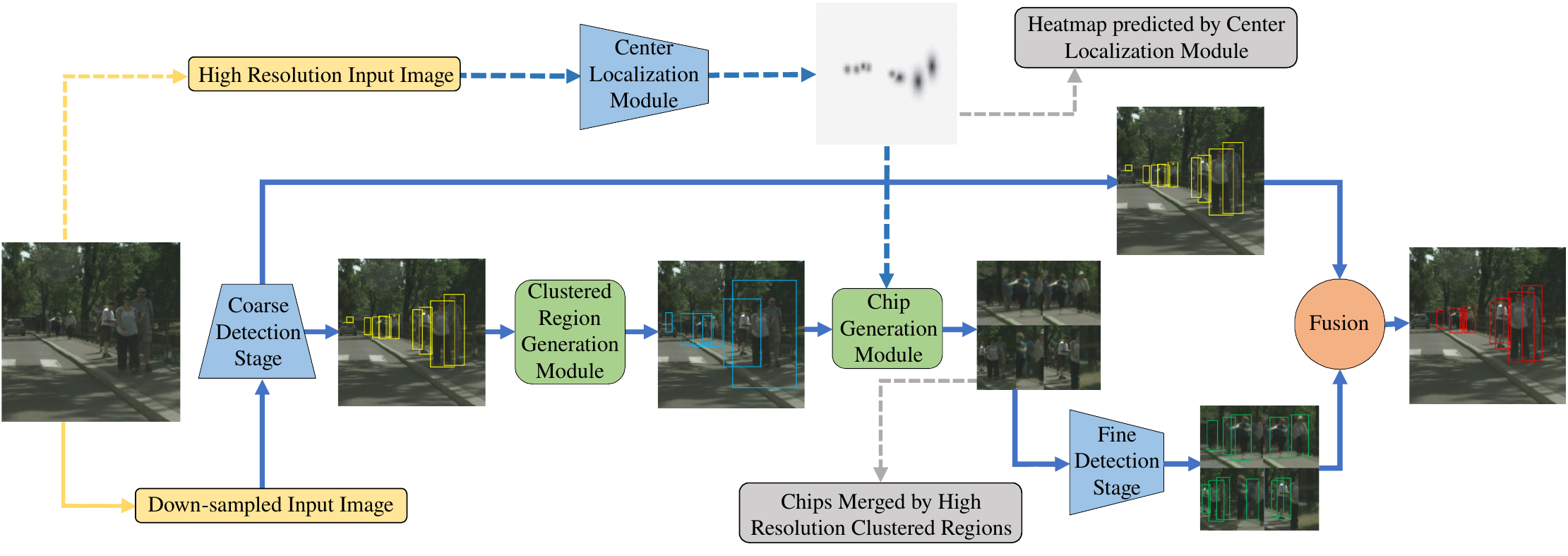}
	\end{minipage}
	\caption{The overall structure of detecting process. Our method consists of five key components: (1) Center localization for small object (2) Coarse detection on the down-sampled high-resolution image; (3) a cluster region generation module to generate cluster regions; (4) a chip generation module to generate chips from the cluster regions; and (5) Fine detection on  chips. The final results are generated by fusing detection results from chips and global image. For simplify, we only demonstrate a small region of an large image in this figure. The yellow boxes in image represent the coarse detection results. The blue boxes in image represent the cluster region boxes. The green boxes in image are the fine detection results and the red boxes are the final results for the original image.}
	\label{fig:2}
\end{figure*}

As shown in Figure \ref{fig:2}, our approach consists of five components, which are center localization for small objects, coarse detection, cluster region and chip generation, fine detection on chips and fusion of results. In specific, after the coarse detection of original down-sampled image and obtains center locations of small objects in high-resolution image by lightweight detector at first-stage, our approach takes the processed results and outputs the cluster regions. In order to avoid processing too many chips, we propose an resize and splice module to reduce the chip number. Then, the chips are fed to fine processing stage for accurately detection. The final detection is obtained by fusing the detection results of each cluster chip and global down-sampled image with non-maximum suppression (NMS).

\subsection{Center Localization for Small Objects}
\subsubsection{Background for Center Localization}
The anchor-free methods of object detection can be divided into three parts: center localization, object classification, and size regression. Although small objects can be better detected by directly inputting high-resolution images to detector, it is difficult to converge parameters using large networks because of small mini-batch size, and the detection accuracy decreases when switching to lightweight networks because of the limited representation capacity. We observe that the small objects can still be satisfactorily  detected with lightweight detection network by high-resolution input image. Therefore, we can predict the center location to locate small objects (smaller than 96×96 pixels) as much as possible in high-resolution image by using a lightweight network. And we only keep center localization and neglect classification and size regression of small objects in high-resolution image to simplify task. Only predict the heat-map without classification information is more accurate and less computationally than that with classification and size regression simultaneously. Inspired by TTFNet\cite{ttfnet}, we adopt the Gaussian kernel as in CornerNet\cite{cornernet} to produce a heat-map and predict the center location of objects by the heat-map. The Gaussian kernel enables the network to produce higher activations near the object center\cite{ttfnet}. And our approach does not require any other predictions to help correct the error as in TTFNet\cite{ttfnet}. Then the interest regions are extracted for further processing according to the predicted center points. 

\begin{figure}[htb]
	\begin{minipage}[b]{1.0\linewidth}
		\centering
		\includegraphics[width=8.5cm]{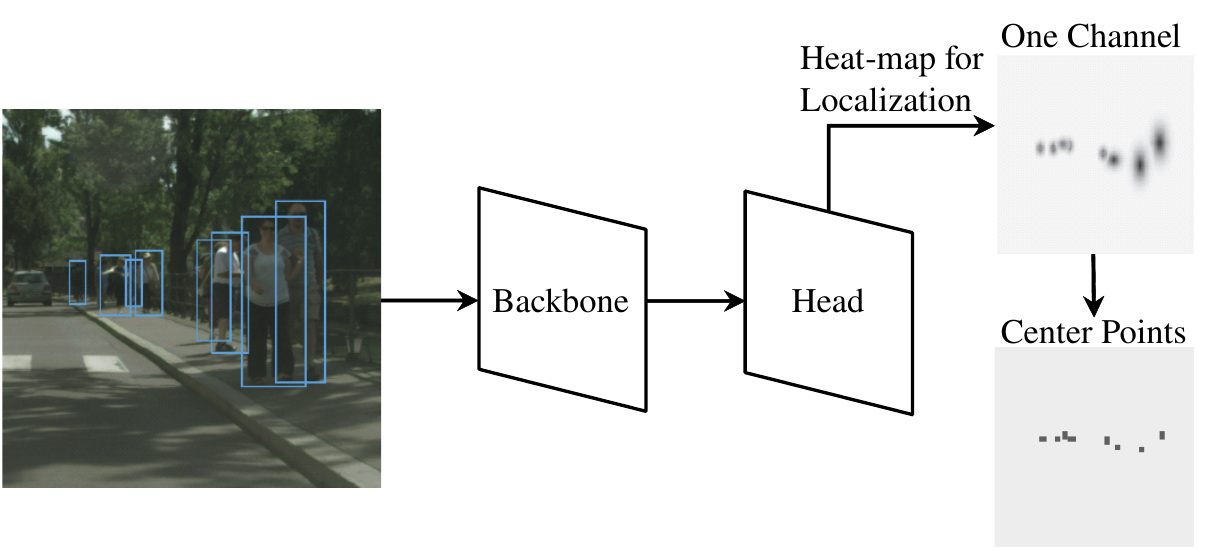}
	\end{minipage}
	\caption{Architecture and pipeline of center localization module. Features are extracted by a backbone network and then down-sampled to 1/4 resolution of the original image. Then, the features are used for localization task. For localization, the network can produce higher activations near the object center.}
	\label{fig:3}
\end{figure}

\subsubsection{Object Center Localization}
Given an image, our network outputs $\hat{H}\in R^{1\times \frac{H_r}{r}\times \frac{W_r}{r}}$. The feature $\hat{H}$, also called heat-map, contains the center location information of object. $H_r$, $W_r$, $r$ are height and width of the input image, and output down-sample stride from the input image size to the heat-map size, respectively. We set $r=4$ in experiments. 
 
we use $M$ to denote the number of the objects and $m$ to denote the $m$-th object's annotated bounding box in an image. The output heat-map is made up of the linearly mapped feature-map scale of each annotated bounding box in the input image. The 2D gaussian kernel $\mathrm{K}_m\left( x,y \right) =\exp \left( -\frac{\left( x-x_0 \right) ^2}{2\sigma _{x}^{2}}-\frac{\left( y-y_0 \right) ^2}{2\sigma _{y}^{2}} \right)$ is adopted to produce $H_m\in R^{1\times \frac{H}{r}\times \frac{W}{r}}$, where $\sigma _x=\frac{\beta w_b}{6}$, $\sigma _y=\frac{\beta h_b}{6}$, $\left( x_0,y_0 \right) _m$ is center location and $(w_b,h_b)$ is the width and height of the annotated bounding box. We define scaler $\beta=0.54$ as in TTFNet\cite{ttfnet} to control the kernel size. Then, we update the output heat-map $H$ by applying element-wise maximum with $H_m$. We use $\left( \lfloor \frac{x}{r} \rfloor ,\lfloor \frac{y}{r} \rfloor \right)$ to force the center to be in the pixel as in CenterNet.

The peak of the Gaussian distribution is the same as the pixel of the box center, which is treated as the positive sample while any other pixel is treated as the negative sample. The loss function is the same as in TTFNet except without the classification channel $c$. Given the $\hat{H}$ and the ground truth output heat-map target $H$, we have 
\begin{equation}
L=\frac{1}{M}\sum_{xy}^{}{\begin{cases}
		\left( 1-\hat{H}_{ij} \right) ^{\alpha _f}\log \left( \hat{H}_{ij} \right) \,\,\,\,\, \mathrm{if} H_{ij}=1\\
		\left( 1-\hat{H}_{ij} \right) ^{\beta _f}\hat{H}_{ij}^{\alpha _f}\log \left( 1-\hat{H}_{ij} \right) \,\, \mathrm{else}\\
\end{cases}}
\end{equation}
where $\alpha _f$ and $\beta _f$ are hyper-parameters in focal loss\cite{retinanet} and its modified version\cite{cornernet, centernet}, respectively. $M$ represents the number of annotated boxes and $L$ is the total loss. We set $\alpha _f=2$ and $\beta _f=4$.

\subsubsection{Overall Design of Center Localization Module}
The architecture of center localization module is shown in Figure \ref{fig:3}. The backbone can use any state-of-the-art structure. Here we use ResNet18\cite{resnet} as the backbone in our experiments. The features extracted by the backbone are down-sampled to 1/4 resolution of the input image, which is implemented by Modulated Deformable Convolution (MDCN)\cite{dconv} and up-sample layer. MDCN layers are followed by Batch Normalization (BN) and ReLU. 

The down-sampled features then go through the head to obtain the localization heat-map that produces high activations on positions near the object center. Since the object center is the local maximum at the heat-map, we can suppress local non-maximum values with the help of 2D max-pooling as in TTFNet\cite{ttfnet}. Then we can attain the object center results. 

To improve the detection performance on small objects, we add the shortcut connections from low-level features to high-level features. The shortcut connections introduce the features from stage 2, 3, and 4 of the backbone, and each connection is implemented by 3 × 3 convolution layers. The number of the layers are set to 3, 2, and 1 for stages 2, 3, and 4, and ReLU follows each layer except for the last one in the shortcut connection.

\begin{figure}[htb]
	\begin{minipage}[b]{1.0\linewidth}
		\centering
		\includegraphics[width=6.5cm]{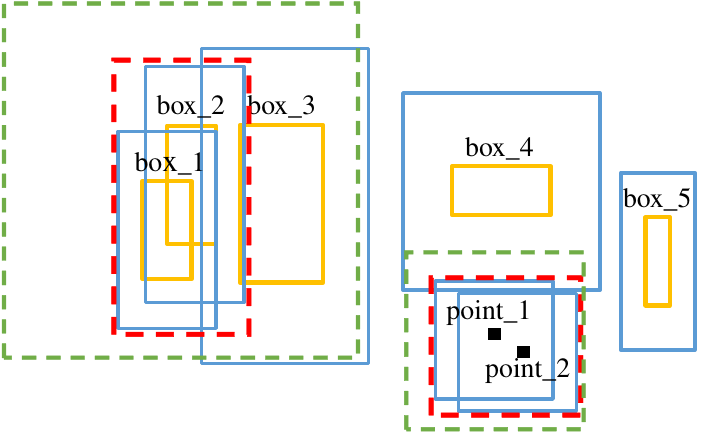}
	\end{minipage}
	\caption{A sample of cluster region boxes generation process. The yellow boxes are the detected boxes of coarse detection. The black points are center localization results. The blue boxes are the expanded coarse proposal boxes. The height and width of green box is the limit size of cluster region box based on `box\_2'. The red box is the final cluster region box. The blue boxes' overlap of `box\_1' and `box\_2' is larger than the threshold $\tau$ and the union is in the green box, so we merge these blue boxes to red box as an cluster region box. And the same as the `point\_1' and `point\_2'.}
	\label{fig:4}
\end{figure}

\subsection{Cluster Region and Chip Generation}
\subsubsection{Generation of Coarse Proposal boxes}
Coarse proposal boxes are generated by the results of coarse detection and center localization stage. Coarse detection can be done using any kind of state-of-the-art detector. Since high-resolution images can cause the GPU out of memory, the detection process can be smoothly carried out by proper down-sampling of the original image. 
Since each detected box has a confidence value, the detected boxes whose confidence value is higher than the set threshold is regarded as the detection results, so the threshold is appropriately set here to reduce the number of false alarm results while extracting regions that contain objects as far as possible. And The number of detected boxes that the confidence value above the threshold is not limited. 

The center localization results are used to supplement the absence of coarse detection results. If the distance of a center localization result and a center point of a coarse detection result is less than 20 pixels, the center localization result is deleted. For the rest of the predicted center localization results, since the width and height of small objects are not predicted, the shape of the coarse proposal boxes is set as square, and the vertical distance from the center point to the four sides of the square is set as $r_c$. For chip generation detailed later, the cluster regions generated by coarse proposal boxes need to be properly scaled, if the value of $r_c$ is too small, the small objects will be up-sampled too many times which will make the object too blurry and harder to detect. Besides, the background information which is helpful to the detection result may be too little. Although the main purpose of center localization is to obtain the location of small objects, many larger objects will also be detected, so the setting of $r_c$ needs to cover most detected small objects. 

\subsubsection{Cluster Region Generation}
Many approaches\cite{snip,sniper,clusdet,densitymap} use image pyramid method or aggregated the detection results into several large cluster regions that contain many objects and then performed fine detection on each cluster region, which is extremely inefficient and only fit the distribution of the objects is non-uniform in image. To avoid this, we obtain cluster regions in image that contains only one detected object (if the objects are densely distributed, a region may have several objects at the same time) and splice them into several large chips. The process from detected results to chips can be divided into the following steps: (1) expanding coarse proposal boxes in a certain way; (2) merging the expanded coarse proposal boxes to several foreground cluster region boxes and (3) padding and resizing cluster regions cropped by cluster region boxes and pack cluster regions into several unified chips.
 
\renewcommand{\algorithmicrequire}{ \textbf{Input:}} 
\renewcommand{\algorithmicensure}{ \textbf{Output:}} 
\begin{algorithm}[H]
	\caption{The Modified Non Max Merge}\label{alg:alg1}
	\begin{algorithmic}
		\REQUIRE Sorted boxes $\mathcal{B} = \left\lbrace B_i\right\rbrace _{i=1}^{N_b} $ in ascending order according to the size of the expanded coarse proposal boxes, height $h^{'}_i$ and width $w^{'}_i$ for $B_i$, height $h_c$ and width $w_c$ for cluster region box $C_k$, the expand ratio $\gamma=1.5$ use for calculate the limited size $(h_r,w_r)$ of $C_k$, $h_{min}$ also limit the size of $C_k$, IoU threshold is $\tau=0.2$.
		\ENSURE The set of cluster region boxes $\mathcal{C} = \left\lbrace C_k\right\rbrace _{k=1}^{N_c}$
		\STATE $\mathcal{C} = \left\lbrace \right\rbrace$
		\STATE \textbf{for} $i \longleftarrow 1$ to $N_b$ \textbf{do}
		\STATE \qquad \textbf{if} $B_i$ is \textbf{visited}
		\STATE \qquad\qquad \textbf{continue}
		\STATE \qquad Flag $B_i$ as \textbf{visited}
		\STATE \qquad $C_k \longleftarrow B_i$
		\STATE \qquad $(h_r,w_r)=(\gamma h^{'}_i,\gamma h^{'}_i)$
		\STATE \qquad \textbf{for} $j=i+1$ to $N_{b}$ \textbf{do}
		\STATE \qquad\qquad \textbf{if} IoU$(B_j,B_i)>\tau$ \textbf{then}
		\STATE \qquad\qquad\qquad $(C_k^{'},h_c^{'},w_c^{'}) \longleftarrow C_k \cup \{B_j\}$
		\STATE \qquad\qquad\qquad \textbf{if} $h_c^{'}>h_{min}$ \textbf{or} $w_c^{'}>h_{min}$ \textbf{then}
		\STATE \qquad\qquad\qquad\qquad \textbf{if} $h_c^{'}>h_r$ \textbf{or} $w_c^{'}>h_r$ \textbf{then}
		\STATE \qquad\qquad\qquad\qquad\qquad \textbf{continue}
		\STATE \qquad\qquad\qquad Flag $B_j$ as \textbf{visited}
		\STATE \qquad\qquad\qquad $C_k \longleftarrow C_k^{'}$
		\STATE \qquad $\mathcal{C} \longleftarrow \mathcal{C} \cup \left\lbrace C_k \right\rbrace$
		\STATE \textbf{return} $\mathcal{C}$
		\label{code:recentEnd}
	\end{algorithmic}
\end{algorithm}

To keep the ground truth boxes of objects in the coarse proposal bounding box as much as possible, we first expand the width and height of each coarse proposal box from the center with an expansion coefficient $\alpha$. Influenced by the chip generation section detailed later, to prevent the resized cluster region from taking up too much area in a chip, for the expansion in the direction of height $h$ and width $w$ of a coarse proposal box, we have
\begin{equation}
\begin{cases}
	h^{'}=\alpha h, w^{'}=1.5\alpha w, \quad if \ h/w>1.5  \\
	h^{'}=\alpha h, w^{'}=\alpha h, \quad if \ h/w<0.75 \\
	h^{'}=\alpha h, w^{'}=\alpha w, \quad others
\end{cases}
\end{equation}
$h^{'}$ and $w^{'}$ represent the height and width of the expanded coarse proposal boxes.

Then we detail the high-quality foreground clusters generation method. In the actual situation, the number of regions generated by an image can be non-identical because of the different object numbers and distribution in different images. The modified Non-maximum Merge (NMM) algorithm proposed by us not only adaptively generates cluster region boxes to reduce the number of coarse proposal boxes but also restricts the size of cluster regions to avoid making large-scale cluster regions that contain too many objects when objects in image are crowded. A sample of cluster region boxes generation process is shown in Figure \ref{fig:4} and we detailed the modified NMM process in Algorithm \ref{alg:alg1}.

In Algorithm \ref{alg:alg1}, $\mathcal{C} = \left\lbrace C_k\right\rbrace _{k=1}^{N_c}$ represents a set $\mathcal{C}$ that have $N_c$ cluster region boxes in an image. $\mathcal{B} = \left\lbrace B_i\right\rbrace _{i=1}^{N_b}$ represents a set $\mathcal{B}$ that have $N_b$ expanded coarse proposal boxes. All boxes in $\mathcal{B}$ are sorted in ascending order according to the box size and starting from the smallest expanded box $B_i$ with size of $(w^{'}_i, h^{'}_i)$. Then we select other boxes in $\mathcal{B}$ whose IoU with $B_i$ are larger than the IoU threshold $\tau$. We don't limit the the size of $C_k$ if the height or width of $C_k$ less than $h_{min}$. But if the  the height or width of $C_k$ larger than $h_{min}$, we limit the size of $C_k$ less than $(h_r, w_r)$. $(h_r, w_r)$ is set to $(\gamma h^{'}_i, \gamma h^{'}_i)$ and we set $\gamma=1.5$. When The final position and size of $C_k$ is a box that contains all the selected boxes in $\mathcal{B}$ and has the minimum size of the box. Therefore, when a selected box causes the size of $C_k$ out of the limit above, the box will not be selected in $C_k$. All the boxes selected in $C_k$ by IoU threshold are set visited and no longer participate in the subsequent cluster region generate process. Afterwards, we repeat the aforementioned process until $\mathcal{B}$ is empty. Note that some objects may be cropped into two parts during the merge process. Therefore, we set up the IoU threshold $\tau=0.2$ to allow a slight overlap among the boxes.

\begin{figure}[htb]
	\begin{minipage}[b]{1.0\linewidth}
		\centering
		\includegraphics[width=6.5cm]{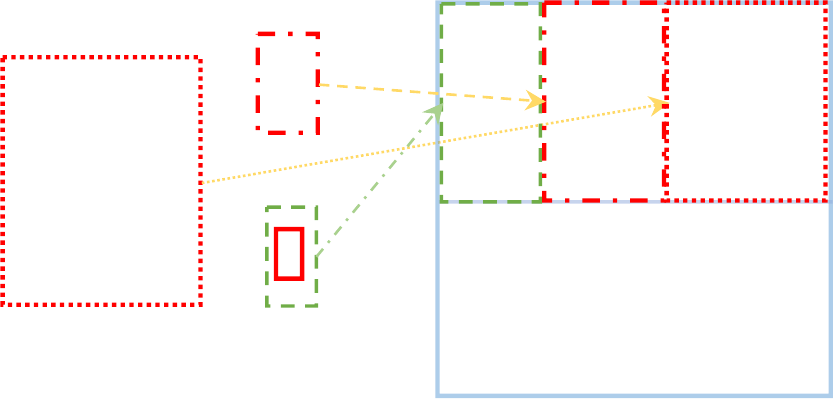}
	\end{minipage}
	\caption{A sample of chip generation process. Red boxes on the left represent the cluster regions. The green box on the left which the height is $h_{min}$ is the expanded cluster region box. The blue box which height and width is $2H$ ($a=2$) is the chip. We resize the regions on the left and splice the regions in blue chips as shown on the right. The blue chip is the image data for fine detection.}
	\label{fig:5}
\end{figure}

\subsubsection{Chip Generation}
Then we pad and resize cluster regions cropped by cluster region boxes and pack cluster regions into several unified chips. Since the width $w_c$ and height $h_c$ of each cluster region are different, to prevent the small regions from being up-sampled too many times, which will make the region too blurry and hard to the detect object in it, we expand the height of the cluster region box less than $h_{min}$ to $h_{min}$ and the width to $w_c h_c/h_{min}$. The $h_{min}$ is the same as the parameter in the modified NMM algorithm.

\begin{figure}[htb]
	\begin{minipage}[b]{1.0\linewidth}
		\centering
		\includegraphics[width=8.5cm]{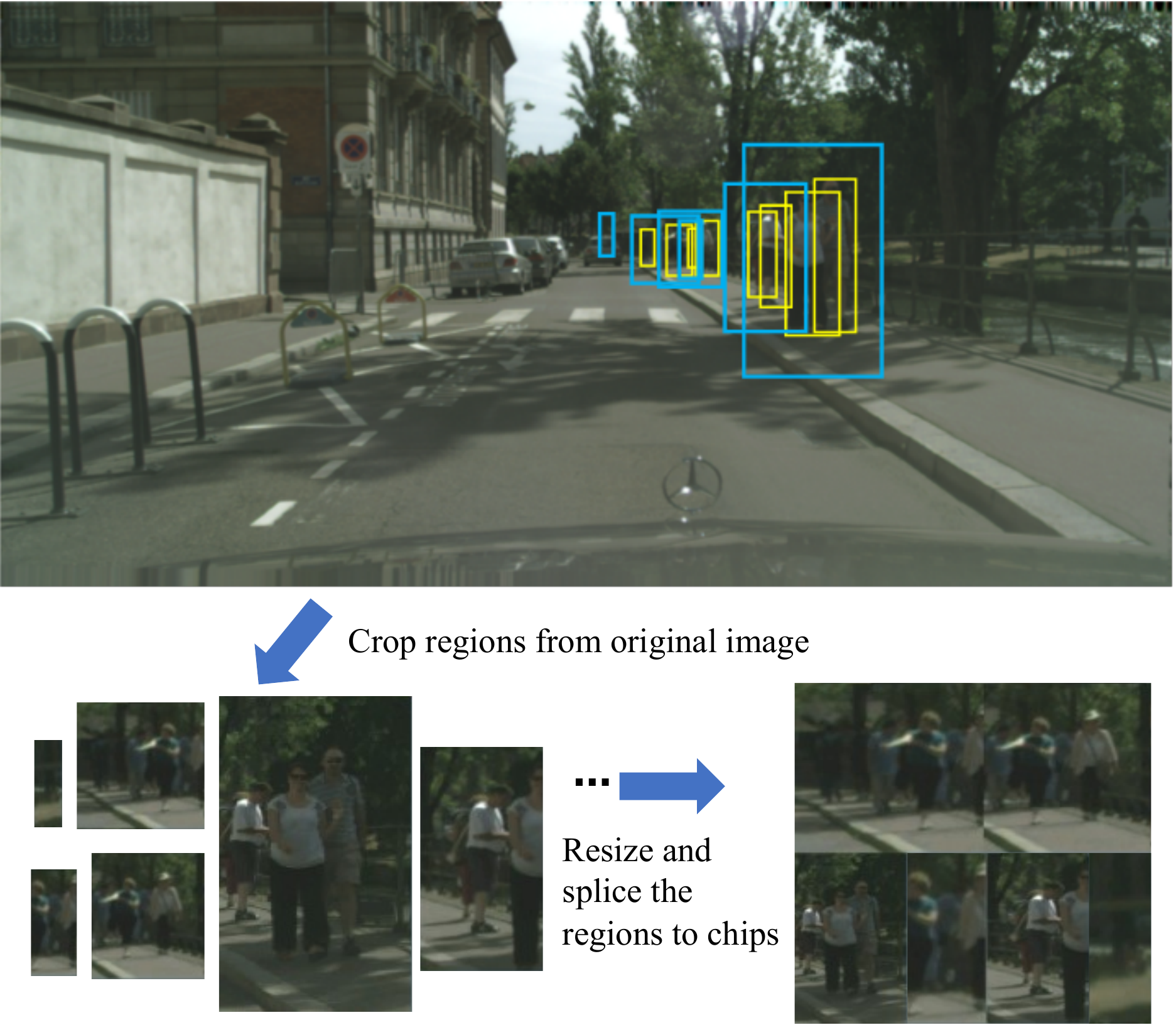}
	\end{minipage}
	\caption{There may be many cluster regions, to reduce the cluster region number, we resized and spliced them into chips. Yellow boxes are original detect results, and blue boxes are cluster region boxes generated by the modified NMM algorithm. We first roughly locate objects, and then accurately detect them via high-resolution detailed chips. This avoids the problem that directly feeding high-resolution images to the detector and causing the detector to be too overloaded and inefficient.}
	\label{fig:6}
\end{figure}

After expanding the size of the small cluster region boxes, all these cluster regions are cropped from the original resolution image to make full use of the high-resolution information in image. To facilitate the splicing and improve the resolution of small-scale cluster regions, we fix the height of each resized region as $H_r$, and then scale the width to $w_c H_r/h_c$. If the height of a cluster region is larger than $H_r$, the objects in this region will be down-sampled. Then a chip is obtained by splicing the cluster regions into a larger image like a mosaic. The general process is shown in Figure \ref{fig:6} and a sample of a chip generation process is shown in Figure \ref{fig:5}. We set the height of a chip as $aH_r$, $a$ and $H_r$ are the number of rows and the height of the resized cluster region, respectively. The width of a chip is the same as the height. Since the size of the a chip is fixed, and the number of cluster regions cropped from each image is uncertain, if a chip is already filled, another chip will be added to splice the rest of cluster regions. Each chip is then fed into the detector for fine detection. Therefore, the height's up-sample times of all the cluster regions are no more than $H_r/h_{min}$.

\subsection{Fine Detection Process and Results Fusion}
Fine detection process can also be conducted by any state-of-the-art detector. Each chip is detected in turn and obtains detection results. Since the positions of extracted cluster regions in chip are different from those in the original image, detection results of each cluster region in chip need to be mapped back to the positions in original image. Final outputs are generated by merging the coarse detection results and results in chips via non-maximum suppression.

\begin{figure}[htb]
	\begin{minipage}[b]{1.0\linewidth}
		\centering
		\includegraphics[width=7.5cm]{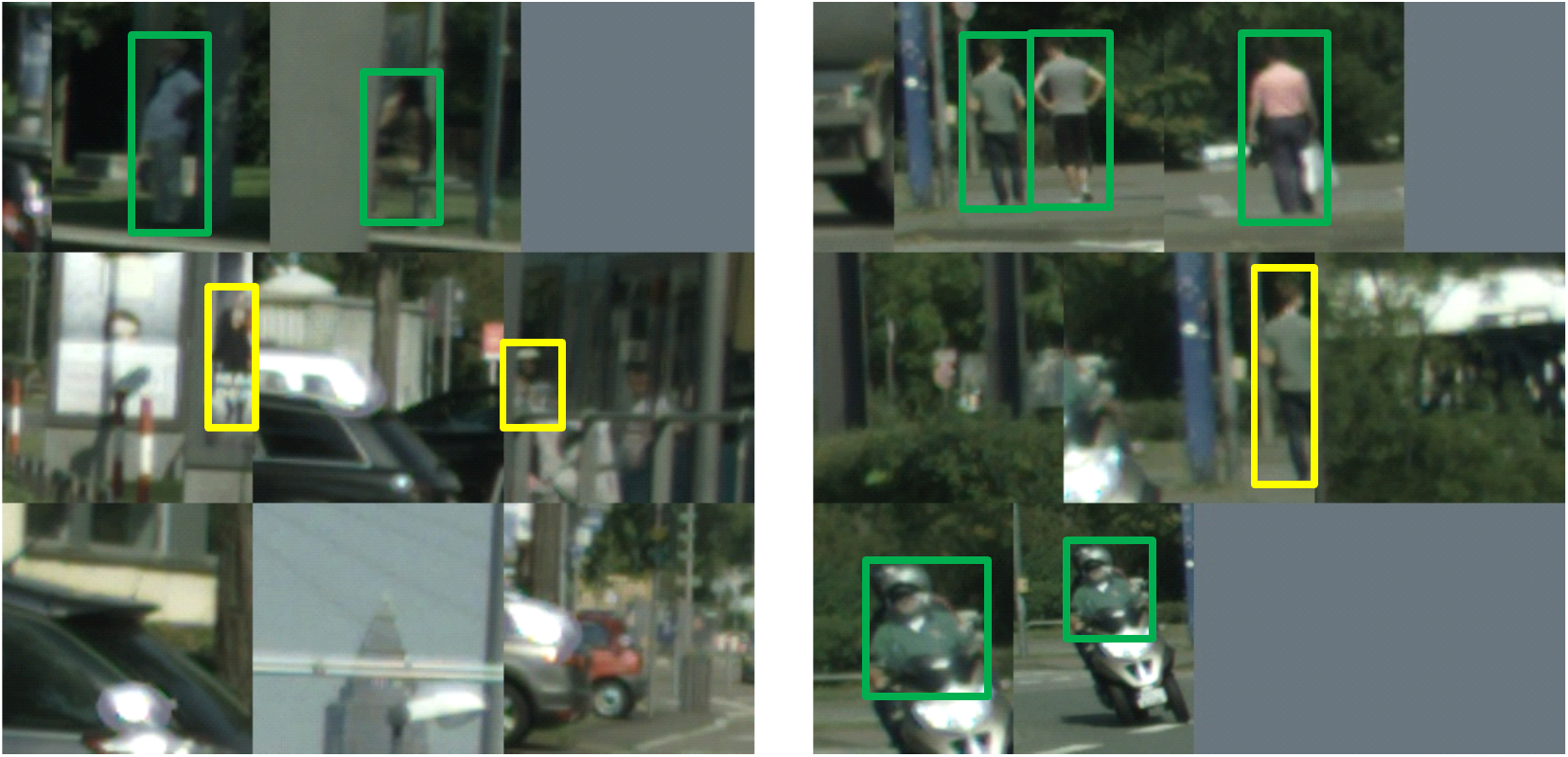}
	\end{minipage}
	\caption{Filter out the result boxes that are obviously incorrect. The green boxes in chips are reserved and the yellow ones are unreserved. The edge of the box that are not at the boundary of the original image but the edge of the box at the boundry oe seriously exceed the croped cluster region are unreserved.}
	\label{fig:7}
\end{figure}

In some cases, an object may be not complete in a chip but still be detected or a box exceeds the crop region’s boundary. These cases can be filtered out before results fusion, as shown in Figure \ref{fig:7}. The detection results corresponding to large objects in the original image contained in chips are ignored for results fusion.

\subsection{Training and Testing of the Framework}
The training process includes coarse detection, fine detection, and center localization for small objects. The train of center localization is to take high-resolution images as input, train with a lightweight network, and predict the heat-map of the object distribution on the training dataset and testing dataset respectively, then obtain the location of the center point and get the possible region of the object. Center localization for small objects is trained only with objects smaller than 96×96 pixels in the dataset. 

The training process of coarse detection and fine detection can directly use the existing state-of-the-art object detection network. The training process of coarse detection takes down-sampled images as input and then detects the training dataset and test dataset respectively by using the trained weight and obtaining coarse detection results. The training process of fine detection take the chips generated by our approach as input and obtains fine detection results. 

The training dataset of fine detection is generated by merging coarse detection results, ground truth boxes, and center localization results of training dataset. The generation of test dataset is the same as training dataset but without ground truth boxes and the generation process is detailed above. For ground truth labels of fine detection dataset, if the intersection size of a cluster region box and a ground truth box is larger than 0.85 times of this ground truth box's size, we transfer the position of this box to chips.

\section{Experiment}
We conduct experiments to evaluate the effectiveness of our proposed method. Our approach is evaluated on the CityPersons\cite{citypersons} and TinyPerson\cite{tinyperson} benchmarks and extensive experiments are carried out. 
\subsection{Datasets and Evaluate Metrics}
\begin{figure}[htb]
	\begin{minipage}[b]{1.0\linewidth}
		\centering
		\includegraphics[width=9.0cm]{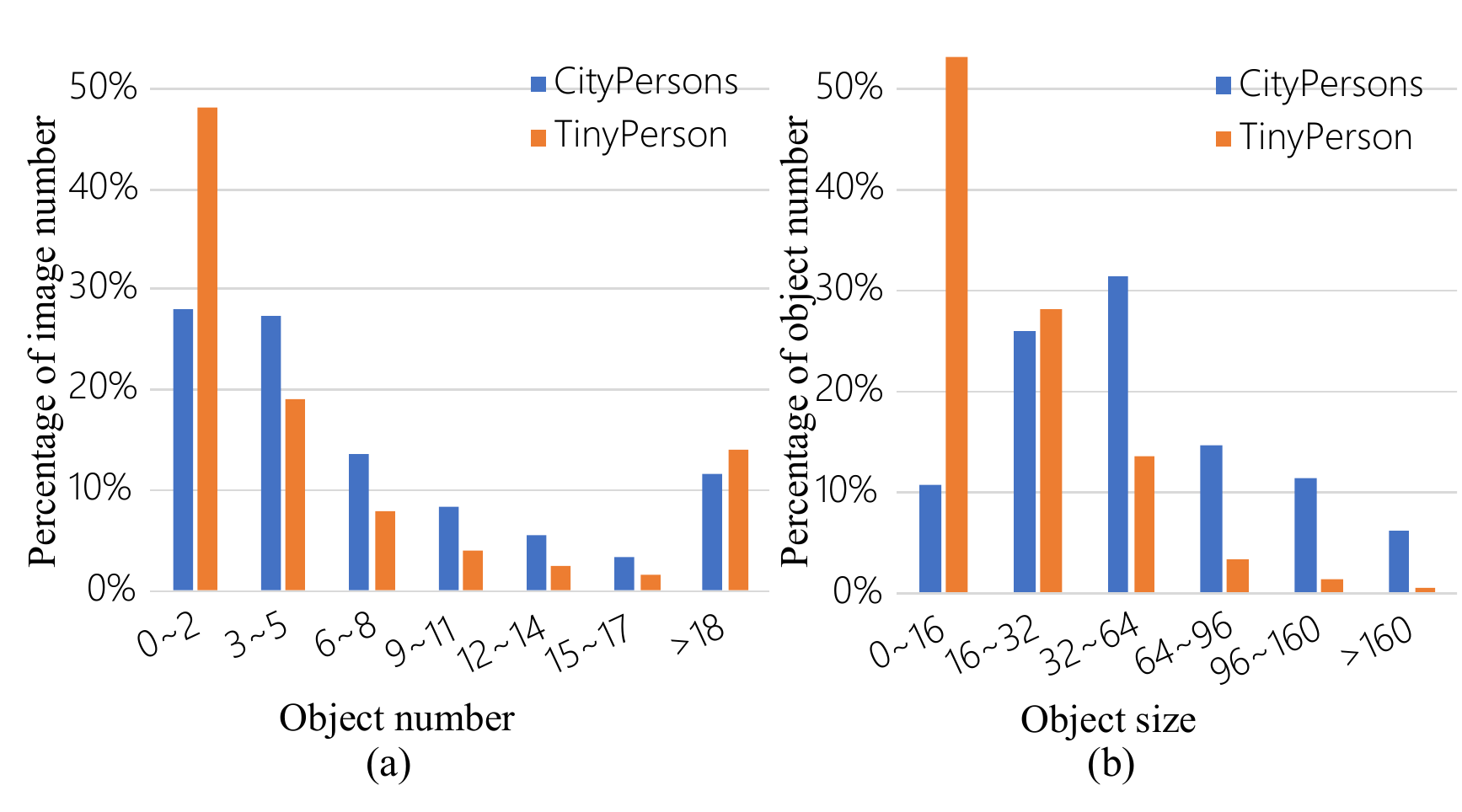}
	\end{minipage}
	\caption{In (a), we statistic the image number through the object number in every image. (b) is the distribution of object size. The X-axis is the square root of the annotated bounding box size. We can notice that the number of objects in most images is small, and the small and medium size of objects are the majority in datasets. The object size is smaller in TinyPerson compared with CityPersons.}
	\label{fig:8}
\end{figure}

CityPersons\cite{citypersons} dataset consists of 2428 training images and 500 validation images with one category, the resolution of the image is 2048×1024. In Figure \ref{fig:8}, we give the statistical information of the object in datasets. It can be seen from Figure \ref{fig:8} that the distribution of objects in most images is sparse, and small and medium objects account for the main part. We use the validation set to evaluate different models because the web test server is shut down and we do not have labels for test set. Since the objects in the dataset may occlude each other, we only select the visible parts of objects as bounding boxes. Some billboards or traffic signs also have human signs but are not real pedestrians, which we remove in the training and testing process. 

TinyPerson\cite{tinyperson} dataset consists of 794 training images and 816 test images with one category, the resolution of most images is 1920×1080 or even higher. The images in the dataset are cut into some sub-images with overlapping during training and test. We choose the sub-images with the size of 1920×1080 as our dataset. The sub-images consist of 851 training images and 2187 test images. Because the number of test images is larger than the number of train images and the number of objects in train dataset is almost the same as in test dataset, We use the train dataset as the test dataset and the test dataset as train dataset. We ignore images with dense objects (more than 200 persons per image) and objects that are defined as ignored and uncertain. In Figure \ref{fig:8}, we also give the statistical information of the object in TinyPerson\cite{tinyperson}. Most objects are small objects in TinyPerson\cite{tinyperson}. 

Similar to the protocols for general object detection (MS COCO\cite{coco} dataset), we adopt Average Precision (AP) and APs at the IoU threshold of 0.5 ($AP_{50}$) and 0.75 ($AP_{75}$) as the metrics on both the datasets. For CityPersons\cite{citypersons} and TinyPerson\cite{tinyperson} dataset, the object size range is divided into 3 intervals by pixel number: small$[0, 32^2]$, medium$[32^2, 96^2]$ and large$[96^2, inf]$. 

We use $G_{ij}=(x_{ij}, y_{ij}, w_{ij}, h_{ij})$ to describe the $j$-th object’s bounding box of $i$-th image $I_i$ in dataset as in \cite{tinyperson}, where $(x_{ij}, y_{ij})$ denotes the coordinate of the left-top point, and $w_{ij}$, $h_{ij}$ are the width and height of the bounding box. $W_i$, $H_i$ denote the width and height of $I_i$, respectively. We define relative size (RS) as $\sqrt{w_{ij} \cdot h_{ij}}$ and absolute size (AS) as $\sqrt{\frac{w_{ij}\cdot h_{ij}}{W_i \cdot H_i}}$. The absolute size of the dataset with high-resolution images may be similar to that of the MS COCO\cite{coco} dataset, but the relative size is much smaller. Although an up-sampled or down-sampled image can change the absolute size, but can't increase the relative size. In theory, increasing the relative size or absolute size appropriately can increase the average precision of object detection. The relative size, absolute size, and aspect ratio of objects in different datasets are shown in Table \ref{tab:table3}. The number on the left side and right side of the `±' is the mean and standard deviation of object size, respectively. The `s', `m', and `l' represent the ratio of the small, medium, and large objects in dataset, respectively.


\subsection{Implementation Details}
For TinyPerson\cite{tinyperson} dataset, we implemented the proposed approach on Pytorch 1.11. The coarse detection and fine detection models are trained with two RTX 2080Ti GPUs. We trained our models for 100 epochs, with the learning rate starting from $1.25\times10^{-4}$ for TTFNet\cite{ttfnet}, decaying by 10 times at 60 and 80, respectively. For coarse detection stage, excessive down-sampling of input image will lead to non-convergence of training process, while high-resolution of input image will lead to out of memory or slow convergence due to small batch size. Therefore, the input image size is 1333×800, and the batch size is 6. The coarse detection threshold is set to 0.10. For fine detection stage, the input size is 640×640 and the batch size is 8. The detection threshold for fine detection is set to 0.10. For chip generation, there are four lines of regions in one chip. The height and width of chip are 640×640 and the height $H$ of resized cluster region is 160. The merge threshold is set to 0.3 in the modified NMM. The box expand ratio $\alpha$ and $h_{min}$ is 3.0 and 80, respectively. For results fusion stage, The IoU threshold of the NMS is set to 0.6. For the center localization for small objects, the size of the input image is 1920×1080, the center point threshold is set as 0.10, and the vertical distance from the center point to the edge of the rectangle is set as 48 determined by the aspect ratio of dataset. To prevent the detection effect from being reduced, results fusion stage is not used because most objects are small objects in TinyPerson and the AP of coarse detection stage is very low. The input size of detector is the same as that in training phase whenever not specified. 

For CityPersons\cite{citypersons}, the input size is 480×480 and the batch size is 12 for fine detection stage. For chip generation, there are three lines of regions in one chip. The height and width of chip are 480×480 and the height $H_r$ of resized cluster region is 160. The coarse detection threshold is set to 0.10. For the center localization for small objects, the size of the input image is 2048×1024, the center point threshold is set as 0.10. Results fusion stage is used for CityPersons to improve the accuracy of large objects. The other parameters are the same as in TinyPerson.

\begin{table*} 
	\centering
	\caption{The detection performance on Citypersons dataset.\label{tab:table1}}
	\begin{tabular}{cccccccccc}
		\hline
		Methods & s1 & s2 & \#img & $AP$ & $AP_{50}$ & $AP_{75}$ & $AP_{s}$ & $AP_m$ & $AP_l$ \\
		\hline
		FRCNN\cite{frcnn} & 768*384 & & 500 & 16.4 & 36.0 & 13.4 & 0.3 & 15.7 & 50.0    \\
		FRCNN\cite{frcnn} & 1333*800 & & 500 & 24.7 & 50.6 & 22.2 & 2.2 & 29.4 & 57.6    \\
		FRCNN\cite{frcnn}+ours & 768*384 & 480*480 & 1601 & 27.0 & 58.5 & 21.2 & 11.8 & 29.7 & 45.1    \\
		FRCNN\cite{frcnn}+ours & 1333*800 & 480*480 & 1515 & 30.9 & 62.0 & 27.6 & 10.8 & 34.9 & 54.1    \\
		RetinaNet\cite{retinanet} &768*384 & & 500 & 17.6 & 38.0 & 14.3 & 0.8 & 19.1 & 50.4   \\
		RetinaNet\cite{retinanet} &1333*800 & & 500 & 27.1 & 54.1 & 24.7 & 2.4 & 35.0 & 55.8   \\
		RetinaNet\cite{retinanet}+OIP & & 768*768 & 3000 & 26.8 & 58.6 & 21.0 & 6.6 & 34.2 & 50.6   \\
		RetinaNet\cite{retinanet}+ours &1333*800 & 480*480 & 1546 & 31.2 & 63.0 & 27.7 & 9.4 & 36.0 & 55.5   \\
		TTFNet\cite{ttfnet}  &768*384 & & 500 & 27.2 & 54.0 & 25.5 & 5.4 & 33.5 & 57.7  \\
		TTFNet\cite{ttfnet}  &1333*800 & & 500 & 38.9 & 69.5 & 40.3 & 14.1 & 44.7 & 64.0  \\
		TTFNet\cite{ttfnet}+OIP  & & 768*768 & 3000 & 43.3 & 73.6 & 45.3 & 18.2 & 49.6 & \textbf{65.2}  \\
		\hline
		TTFNet\cite{ttfnet}+ours  &768*384 & 480*480 & 1668 & 41.2 & 72.6 & 41.9 & 22.2 & 45.4 & 63.5  \\
		TTFNet\cite{ttfnet}+ours  &1333*800 & 480*480 & 1668 & \textbf{44.4} & \textbf{75.1} & \textbf{47.9} & \textbf{21.7} & \textbf{50.5} & 64.5  \\
		\hline
	\end{tabular}
	\label{tab:booktabs}
\end{table*}

\begin{table*} 
	\centering
	\caption{The detection performance on TinyPerson dataset.\label{tab:table2}}
	\begin{tabular}{cccccccccc}
		\hline
		Methods & s1 & s2 & \#img & $AP$ & $AP_{50}$ & $AP_{75}$ & $AP_{s}$ & $AP_m$ & $AP_l$ \\
		\hline
		TTFNet\cite{ttfnet}  &768*384 & & 851 & 3.9 & 15.1 & 0.6 & 3.3 & 10.8 & 2.9  \\
		TTFNet\cite{ttfnet}  &1333*800 & & 851 & 9.1 & 30.5 & 2.6 & 9.0 & 12.4 & 3.2  \\
		TTFNet\cite{ttfnet}+OIP  & & 640*640 & 4835 & 13.6 & 41.7 & 5.1 & 13.1 & 17.1 & 9.3  \\
		\hline
		TTFNet\cite{ttfnet}+ours  &768*384 & 640*640 & 3733 & 15.8 & 45.6 & \textbf{6.7} & 15.0 & 19.7 & \textbf{14.5}  \\
		TTFNet\cite{ttfnet}+ours  &1333*800 & 640*640 & 3479 & \textbf{16.4} & \textbf{46.4} & 6.6 & \textbf{15.4} & \textbf{20.2} & 12.7 \\
		\hline
	\end{tabular}
	\label{tab:booktabs}
\end{table*}

\subsection{Experiment Results}
We compare our approach with the overlap image partition (OIP) method to cut high-resolution images into some sub-images with overlapping on all datasets. Compared with evenly image partition\cite{clusdet}, we implement OIP according to the property of the datasets to prevent the same object from being split into two different parts which may have hurt the AP of large objects. For TinyPerson\cite{tinyperson}, the image size after partition is 640×640. For CityPerson\cite{citypersons}, the street scenario has regular distribution of objects, so we set the image size as 768×768 to keep large objects in one image part as much as possible. The parameters for training stage of image partition are the same as the fine detection stage. In addition, we also compare our method with representative state-of-the-art detectors on all datasets, including Faster-RCNN\cite{frcnn}, RetinaNet\cite{retinanet} or TTFNet\cite{ttfnet}. The backbone of our approach based on Faster-RCNN\cite{frcnn} and RetinaNet\cite{retinanet} is ResNet-50\cite{resnet} and the backbone of TTFNet\cite{ttfnet} is DarkNet-53\cite{yolov3}. The coarse detection stage and fine detection stage use the same state-of-the-art detector for our approach. 

As shown in Table \ref{tab:table1} and Table \ref{tab:table2}, The `s', `m', and `l' represent small, medium, and large, respectively. The `s1' and `s2' represent the input image size of coarse detection and fine detection, respectively. The `\#img' is the number of images forwarded to the detector of fine detection and coarse detection. Experiment results show that our proposed approach achieves promising performance compared with other state-of-the-art detectors or methods. Our approach has a significant improvement over the detection with the down-sampled high-resolution image, especially for small and medium objects. The accuracy of large objects has a slight decrease compared with OIP method due to the down-sample in chip and NMS in the result fusion stage of large objects. The OIP method of high-resolution image can increase the detection accuracy, but greatly increase the image number to process. Our approach can reduce the number of images that need to process and improve the detection accuracy at the same time. When using the same backbone, our approach consistently outperforms such counterparts by margins, improving APs of the second-best by 4.4\% and 1.1\% with RetinaNet\cite{retinanet} and TTFNet\cite{ttfnet} in Table \ref{tab:table1} for CityPersons, respectively. For TinyPerson in Table \ref{tab:table2}, our approach largely boosts the performance of other methods, and it improves the $AP$, $AP_{50}$ and $AP_{75}$ of the second-best by 2.8\%, 4.7\% and 1.6\%, respectively. 

As shown in Table \ref{tab:table3}, compared with the original image, the relative size in chip generated by our approach has greatly improved, and the standard deviation of absolute size has decreased. Therefore, our approach improves the detection accuracy of small and medium objects by generating chips, ensures the detection accuracy of large objects through the coarse detection stage, and improves the accuracy of small objects through the center localization stage. 

Therefore, coarse detection is mainly used for detecting large objects and locating the approximate bounding boxes of the small and medium objects, if many small objects in the image are down-sampled, the precision of small objects can be seriously reduced, the center location is mainly used for locating the small object and reducing false alarm results, the fine detection is mainly used for improving the detection accuracy of the medium and small object. 

\begin{table*}
	\centering
	\caption{Mean and standard deviation of absolute size, relative size and aspect ratio of the datasets.\label{tab:table3}}
	\begin{tabular}{ccccccc}
		\hline
		dataset & absolute size & relative size & aspect ratio & s & m & l \\
		\hline
		COCO\cite{coco} & 99.5±107.5 & 0.190±0.203 & 1.214±1.339 & 41\% & 34\% & 24\% \\
		TinyPerson\cite{tinyperson} & 22.4±22.7 & 0.016±0.016 & 0.75±0.49 & 81.4\% & 16.8\% & 1.8\% \\
		CityPersons\cite{citypersons} & 79.8±67.5  & 0.055±0.046 & 0.410±0.008 & 36.6\% & 45.9\% & 17.5\% \\
		TinyPerson\_c & 29.9±20.8 & 0.047±0.033 & 0.73±0.45 & 67.0\% & 31.2\% & 1.8\% \\
		CityPersons\_c & 50.1±21.3 & 0.104±0.044 & 0.47±0.26 & 21.7\% & 75.8\% & 2.5\% \\
		\hline
	\end{tabular}
\end{table*}

\subsection{Ablation Study}
We detailly validate the major components, i.e., coarse detection and results fusion, as well as several hyper-parameters in our approach. 
\subsubsection{On each component}
We first validate the effect of different components in our approach with TTFNet\cite{ttfnet} in terms of $AP/AP_{50}/AP_{75}$ (\%). Our approach has three major components: coarse detection, fine detection, and center localization. As shown in Table \ref{tab:table4} and Table \ref{tab:table5}, CD means only use the coarse detection results. FD means only generating chips by coarse detection results but not merging coarse detection results. FD+CL means generating chips by coarse detection and center localization results but not merging coarse detection results. CD+FD means only generating chips by coarse detection results and merging coarse detection results. CD+FD+CL means using all components to get the final results. For CityPersons, FD first improves the baseline accuracy ($AP$) by 4.2\% and CL further increases it by 1.1\%, highlighting their credits. For TinyPerson, $AP$ decreases after merging coarse detection results to final results, because there are too many tiny objects which hard to detect for coarse detection.

\begin{table}[!t]
	\caption{Ablation Study on Each Component of CityPersons.\label{tab:table4}}
	\centering
	\begin{tabular}{ccccccc}
		\hline
		method & $AP$ & $AP_{50}$ & $AP_{75}$ \\
		\hline
		CD & 38.9 & 69.5 & 40.3 \\
		FD & 43.1 & 72.8 & 45.9 \\
		CD+FD & 43.3 & 73.8 & 45.9 \\
		FD+CL & 44.3 & 74.5 & 47.9 \\
		CD+FD+CL & 44.4 & 75.1 & 47.9 \\
		\hline
	\end{tabular}
\end{table}

\begin{table}[!t]
	\caption{Ablation Study on Each Component of TinyPerson.\label{tab:table5}}
	\centering
	\begin{tabular}{ccccccc}
		\hline
		method & $AP$ & $AP_{50}$ & $AP_{75}$ \\
		\hline
		CD & 9.1 & 30.5 & 2.6 \\
		FD & 15.5 & 45.4 & 6.0 \\
		CD+FD & 13.5 & 41.9 & 4.4 \\
		FD+CL & 16.4 & 46.4 & 6.6 \\
		CD+FD+CL & 14.3 & 43.2 & 4.9 \\
		\hline
	\end{tabular}
\end{table}

\subsubsection{On hyper-parameters $\alpha$ and $h_{min}$}
As described, $\alpha$ affects the expanded percentage of coarse proposal boxes in the cluster region generation stage, and the box size increases as $\alpha$ increases. $h_{min}$ restricts the highest up-sample times of the cluster regions is no more than $H_r/h_{min}$ and the highest up-sample times decrease as $h_{min}$ increases. 

In Table \ref{tab:table6} and Table \ref{tab:table7}, we report the detection accuracies for different values of $\alpha$ and $h_{min}$. The AS and RS represent the absolute size and relative size of chips for fine detection, respectively. The results in Table \ref{tab:table6} indicate that our approach reaches the highest performance on TinyPerson when $\alpha$=1.5 and $h_{min}=80$. For CityPersons, our approach reaches the highest performance when which $\alpha$=3.0 and $h_{min}=80$. These parameter results therefore used in our experiments. 

The background information increases as $\alpha$ increases and the same as the absolute size and relative size of objects in chips. Too large $\alpha$ will result in lower detection accuracy. But if $\alpha$ is too small, the accuracy will be reduced due to serious damage to background information. For TinyPerson dataset, the absolute size is very small, if the cluster regions are up-sampled to an extremely large scale, some regions in chips will become blurred and the accuracy will decrease because detectors output many false positives on background or local regions of objects.

\begin{table}[!t]
	\caption{The Impact of $\alpha$ and $h_{min}$ on TinyPerson dataset.\label{tab:table6}}
	\centering
	\begin{tabular}{ccccccc}
		\hline
		$\alpha$ & $h_{min}$ & $AP$ & $AP_{50}$ & $AP_{75}$ & AS & RS \\
		\hline
		2.0 & 80 & 14.7 & 43.4 & 5.6 & 33.5±23.3 & 0.052±0.036 \\
		2.5 & 80 & 15.5 & 44.8 & 6.4 & 31.8±22.0 & 0.050±0.034 \\
		3.0 & 80 & 15.8 & 45.6 & 6.7 & 29.9±20.8 & 0.047±0.033 \\
		4.0 & 80 & 11.3 & 36.0 & 3.8 & 25.8±19.3 & 0.040±0.030 \\
		2.5 & 112 & 15.0 & 44.3 & 6.1 & 26.5±20.9 & 0.041±0.033 \\
		2.5 & 48 & 10.9 & 33.8 & 3.9 & 38.4±22.5 & 0.060±0.035 \\
		\hline
	\end{tabular}
\end{table}

\begin{table}[!t]
	\caption{The Impact of $\alpha$ and $h_{min}$ on CityPersons dataset.\label{tab:table7}}
	\centering
	\begin{tabular}{ccccccc}
		\hline
		$\alpha$ & $h_{min}$ & $AP$ & $AP_{50}$ & $AP_{75}$ & AS & RS \\
		\hline
		1.5 & 80 & 44.4 & 75.1 & 47.9 & 50.1±21.3 & 0.104±0.044 \\
		1.25 & 80 & 44.0 & 74.5 & 46.1 & 55.3±22.8 & 0.115±0.047 \\
		1.5 & 48 & 43.5 & 73.7 & 46.2 & 53.8±21.5 & 0.112±0.045 \\
		2.0 & 80 & 43.4 & 74.5 & 46.3 & 43.2±20.9 & 0.090±0.044 \\
		1.5 & 112 & 43.5 & 73.8 & 46.0 & 47.5±22.6 & 0.099±0.047 \\
		\hline
	\end{tabular}
\end{table}

\subsubsection{On input image size}
We only consider the input image size of coarse detection and fine detection stage. For coarse detection stage, input original image cause out-of-memory, the absolute size decreases as the increase of the down-sampling degree, and the relative size doesn't change. As shown in Table \ref{tab:table1} and Table \ref{tab:table2}, the `s1' represent the input image size of coarse detection and the input size of coarse detection is 1333×800 or 768×384. `\#img2' represents the test chip number for fine detection. 

Reducing the resolution of the input image significantly reduces the accuracy of the state-of-the-art detector's result. However, by using our method, the lower input size of coarse detection stage also can achieve relatively high accuracy for final results, and improving the input size of coarse detection will have a smaller improvement on the final result. Besides, even if coarse detection uses a lower-resolution input image, our approach can still make the average precision higher than that directly under a higher resolution as input. Furthermore, when the coarse detection and benchmark use the same resolution input image, our approach can greatly improve the detection accuracy. 

For fine detection stage, $a$ and $H_r$ represent the number of rows and the height of the resized cluster region as detailed in chip generation. Here we change the $a$ to generate different sizes of chips. As shown in Table \ref{tab:table8}, the number of chip decreases and the input size of chip increases as $a$ increases. The average precision is almost unchanged when $a<6$. Because of out of memory, the batch size is set 6 when $a=6$, the AP has a significant decrease.

\begin{table}[!t]
	\caption{Ablation Study on Input Image Size.\label{tab:table8}}
	\centering
	\begin{tabular}{ccccccc}
		\hline
		a & s2 & \#img2 & batch size & $AP$ & $AP_{50}$ & $AP_{75}$ \\
		\hline
		3 & 480*480 & 4515 & 8 & 15.3 & 44.7 & 6.1 \\
		4 & 640*640 & 2882 & 8 & 15.0 & 44.3 & 6.1 \\
		5 & 800*800 & 1876 & 8 & 15.3 & 44.7 & 6.1 \\
		6 & 960*960 & 1458 & 6 & 13.8 & 41.2 & 5.1 \\
		\hline
	\end{tabular}
\end{table}

\subsubsection{On detection threshold}
The confidence threshold of the coarse detection stage and the center localization stage will affect the number of chips and the recall rate of the first stage. If an object is not detected in coarse detection stage, it will not appear in chip, making it impossible to be detected in fine detection stage. 

As shown in Table \ref{tab:table9}, `thres1' and `thres2' represent the detection threshold of coarse detection and center localization, respectively. `\#img2' represents the test chip number for fine detection. The recall rate represents the ratio of the number of detected objects that are labeled as the ground truth in chips after coarse detection and the number of objects in testing dataset. The increased detection threshold will lead to a smaller number of chips, but at the same time, many objects with low confidence will not appear in chip for fine detection, resulting in a lower AP of final result. The lower detection threshold will increase the number of chips, but AP does not increase significantly or even slightly decrease. It may be that the blurry of some regions increased the confidence of some false alarm objects. The proper detection threshold can add many false alarm objects detected by coarse detection to training dataset of fine detection, and improve the ability of fine detection to prevent the confidence of false alarm objects from increasing.

\begin{table}[!t]
	\caption{Ablation Study on Detection Threshold.\label{tab:table9}}
	\centering
	\begin{tabular}{ccccccc}
		\hline
		thres1 & thres2 & \#img2 & recall rate & $AP$ & $AP_{50}$ & $AP_{75}$ \\
		\hline
		0.05 & 0.05 & 1174 & 0.985 & 44.3 & 74.1 & 47.5 \\
		0.1 & 0.1 & 847 & 0.975 & 44.4 & 75.1 & 47.9 \\
		0.2 & 0.2 & 656 & 0.954 & 43.3 & 73.3 & 46.2 \\
		\hline
	\end{tabular}
\end{table}

\section{Conclusion}
In this paper, we presented a novel and efficient coarse-to-fine object detection method for scenarios where there are many small-scale objects, large-scale variance, and sparsely distributed objects in high-resolution images. Images are first down-sampled and carried out coarse detection. Then our method generates chips with higher resolution for further fine detection. The experiments showed that our method achieved better performance and surpassed the most existing state-of-the-art methods. The current highest resolution object detection dataset is around 1920×1080 pixels, but current camera can take images with a resolution of 5760×3240 pixels. In theory, our approach can detect tiny objects from higher resolution images, and ensure the detection accuracy of large-scale objects. If the distribution of the object is dendity in an image, there will be too many chips, which will affect the detection efficiency. For further work, we can make image datasets of the higher original resolution. With the improvement of image resolution, the small object will lose more details after being down-sampled, and the effect of center localization of small objects will become more important. Or we can use super-resolution not up-sample on detected objects and then for fine detection.

\vspace{11pt}

\begin{IEEEbiography}[{\includegraphics[width=1in,height=1.25in,clip,keepaspectratio]{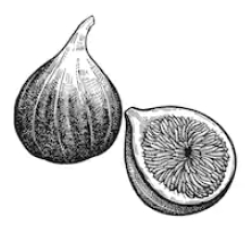}}]{Michael Shell}
Use $\backslash${\tt{begin\{IEEEbiography\}}} and then for the 1st argument use $\backslash${\tt{includegraphics}} to declare and link the author photo.
Use the author name as the 3rd argument followed by the biography text.
\end{IEEEbiography}

\vspace{11pt}

\vfill

\end{document}